\documentclass[article,shortnames]{jss}

\usepackage{graphicx}
\usepackage{amsfonts}
\usepackage{amssymb}
\usepackage{amsmath}

\newcommand{\given}{\operatorname{|}}

\author{Marco Scutari\\University of Padova}
\title{Learning Bayesian Networks with the \pkg{bnlearn} \proglang{R} Package}
\Plainauthor{Marco Scutari}
\Plaintitle{Learning Bayesian Networks with the bnlearn R Package}

\Abstract{

  \pkg{bnlearn} is an \proglang{R} package \citep{r} which includes several algorithms
  for learning the structure of Bayesian networks with either discrete
  or continuous variables. Both constraint-based and score-based algorithms
  are implemented, and can use the functionality provided by the \pkg{snow}
  package \citep{snow} to improve their performance via parallel computing. Several
  network scores and conditional independence algorithms are available
  for both the learning algorithms and independent use. Advanced plotting
  options are provided by the \pkg{Rgraphviz} package \citep{rgraphviz}.

}
\Plainkeywords{bayesian networks, R, structure learning algorithms, constraint-based algorithms,
  score-based algorithms, conditional independence tests}

\Address{
  Marco Scutari\\
  Department of Statistical Sciences\\
  University of Padova\\
  Via Cesare Battisti 241, 35121 Padova, Italy\\
  E-mail: \email{marco.scutari@stat.unipd.it}
}

\begin{document}

\section{Introduction}

In recent years Bayesian networks have been used in many fields, from On-line
Analytical Processing (OLAP) performance enhancement \citep{mphd} to medical
service performance analysis \citep{acid}, gene expression analysis \citep{friedman},
breast cancer prognosis and epidemiology \citep{holmes}.

The high dimensionality of the data sets common in these domains
have led to the development of several learning algorithms focused on reducing
computational complexity while still learning the correct network. Some examples
are the \textit{Grow-Shrink} algorithm in \citet{mphd}, the \textit{Incremental
Association} algorithm and its derivatives in \citet{iamb} and in \citet{fastiamb2},
the \textit{Sparse Candidate} algorithm in \citet{sc}, the Optimal Reinsertion
in \citet{or} and the Greedy Equivalent Search in \citet{ges}.

The aim of the \pkg{bnlearn} package is to provide a free implementation of some
of these structure learning algorithms along with the conditional independence
tests and network scores used to construct the Bayesian network. Both discrete
and continuous data are supported. Furthermore, the learning algorithms can be 
chosen separately from the statistical criterion they are based on (which is usually
not possible in the reference implementation provided by the algorithms' authors),
so that the best combination for the data at hand can be used.

\section{Bayesian networks}

Bayesian networks are graphical models where nodes represent random variables
(the two terms are used interchangeably in this article) and arrows represent
probabilistic dependencies between them \citep{korb}.

The graphical structure $\mathcal{G} = (\mathbf{V}, A)$ of a Bayesian network is a
\textit{directed acyclic graph} (DAG), where $\mathbf{V}$ is the \textit{node}
(or \textit{vertex}) \textit{set} and $A$ is the \textit{arc} (or \textit{edge})
\textit{set}. The DAG defines a factorization of the joint probability distribution
of $\mathbf{V} = \{X_1, X_2, \ldots, X_v\}$, often called the \textit{global
probability distribution}, into a set of \textit{local probability distributions},
one for each variable. The form of the factorization is given by the \textit{Markov
property} of Bayesian networks \citep[section 2.2.4]{korb}, which states that every
random variable $X_i$ directly depends only on its parents $\Pi_{X_i}$:
\begin{align}
  &\Prob(X_1, \ldots, X_v) = \prod_{i=1}^v \Prob(X_i \given \Pi_{X_i})&
  & \text{(for discrete variables)}\\
  &f(X_1, \ldots, X_v) = \prod_{i=1}^v f(X_i \given \Pi_{X_i})&
  & \text{(for continuous variables).}
\end{align}
The correspondence between conditional independence (of the random variables) and
graphical separation (of the corresponding nodes of the graph) has been extended
to an arbitrary triplet of disjoint subsets of $\mathbf{V}$ by \citet{pearl} with
the \textit{d-separation} (from \textit{direction-dependent separation}). 
Therefore model selection algorithms first try to learn the graphical structure
of the Bayesian network (hence the name of \textit{structure learning algorithms})
and then estimate the parameters of the local distribution functions conditional
on the learned structure. This two-step approach has the advantage that it considers
one local distribution function at a time, and it does not require to model the
global distribution function explicitly. Another advantage is that learning algorithms
are able to scale to fit high-dimensional models without incurring in the so-called
\textit{curse of dimensionality}.

Although there are many possible choices for both the global and the local
distribution functions, literature have focused mostly on two cases:
\begin{itemize}
  \item \textit{multinomial data} (the \textit{discrete case}): both the global
    and the local distributions are multinomial, and are represented as probability
    or contingency tables. This is by far the most common assumption, and the
    corresponding Bayesian networks are usually referred to as \textit{discrete
    Bayesian networks} (or simply as \textit{Bayesian networks}).
  \item \textit{multivariate normal data} (the \textit{continuous case}): the
    global distribution is multivariate normal, and the local distributions
    are normal random variables linked by linear constraints. These Bayesian
    networks are called \textit{Gaussian Bayesian networks} in \citet{heckerman3},
    \citet{neapolitan} and most recent literature on the subject.
\end{itemize}
Other distributional assumptions lead to more complex learning algorithms (such
as the nonparametric approach proposed by \citet{mercer}) or present various
limitations due to the difficulty of specifying the distribution functions in
closed form (such as the approach to learn Bayesian network with mixed variables
by \citet{deal}, which does not allow a node associated with a continuous variable
to be the parent of a node associated with a discrete variable).

\section{Structure learning algorithms}

Bayesian network structure learning algorithms can be grouped in two categories:
\begin{itemize}
  \item \textit{constraint-based algorithms}: these algorithms learn the network
    structure by analyzing the probabilistic relations entailed by the Markov
    property of Bayesian networks with conditional independence tests and then
    constructing a graph which satisfies the corresponding d-separation statements.
    The resulting models are often interpreted as \textit{causal models} even
    when learned from observational data \citep{pearl}.
  \item \textit{score-based algorithms}: these algorithms assign a score to each
    candidate Bayesian network and try to maximize it with some heuristic
    search algorithm. Greedy search algorithms (such as \textit{hill-climbing}
    or \textit{tabu search}) are a common choice, but almost any kind of search
    procedure can be used.
\end{itemize}

\textit{Constraint-based algorithms} are all based on the \textit{Inductive
Causation} (IC) algorithm by \citet{ic}, which provides a theoretical framework
for learning the structure causal models. It can be summarized in three steps:
\begin{enumerate}
  \item first the \textit{skeleton} of the network (the undirected graph
    underlying the network structure) is learned. Since an exhaustive search is
    computationally unfeasible for all but the most simple data sets, all
    learning algorithms use some kind of optimization such as restricting
    the search to the \textit{Markov blanket} of each node (which includes
    the parents, the children and all the nodes that share a child with
    that particular node).
  \item set all direction of the arcs that are part of a \textit{v-structure} (a
    triplet of nodes incident on a converging connection $X_j \rightarrow X_i \leftarrow X_k$).
  \item set the directions of the other arcs as needed to satisfy the acyclicity
    constraint.
\end{enumerate}

\textit{Score-based algorithms} on the other hand are simply applications of
various general purpose heuristic search algorithms, such as \textit{hill-climbing},
\textit{tabu search}, \textit{simulated annealing} and various \textit{genetic
algorithms}. The score function is usually \textit{score-equivalent} 
\citep{chickering}, so that networks that define the same probability distribution
are assigned the same score.

\pagebreak

\section{Package implementation}

\subsection{Structure learning algorithms}

\pkg{bnlearn} implements the following constraint-based learning algorithms
(the respective function names are reported in parenthesis):
\begin{itemize}
  \item \emph{Grow-Shrink} (\code{gs}): based on the \emph{Grow-Shrink Markov
    Blanket}, the simplest Markov blanket detection algorithm \citep{mphd} used
    in a structure learning algorithm.
  \item \emph{Incremental Association} (\code{iamb}): based on the Incremental
    Association Markov blanket (IAMB) algorithm \citep{iamb}, which is based on a
    two-phase selection scheme (a forward selection followed by an attempt
    to remove false positives).
  \item \emph{Fast Incremental Association} (\code{fast.iamb}): a variant of IAMB
    which uses speculative stepwise forward selection to reduce the number of
    conditional independence tests \citep{fastiamb2}.
  \item \emph{Interleaved Incremental Association} (\code{inter.iamb}):
    another variant of IAMB which uses forward stepwise selection \citep{iamb}
    to avoid false positives in the Markov blanket detection phase.
  \item \emph{Max-Min Parents and Children} (\code{mmpc}): a
    forward selection technique for neighbourhood detection based on the
    maximization of the minimum association measure observed with any subset
    of the nodes selected in the previous iterations \citep{mmhc}.
    It learns the underlying structure of the Bayesian network (all the arcs
    are undirected, no attempt is made to detect their orientation).
\end{itemize}
Three implementations are provided for each algorithm:
\begin{itemize}
  \item an optimized implementation (used by default) which uses backtracking
    to roughly halve the number of independence tests.
  \item an unoptimized implementation (used when the \code{optimized} argument
    is set to \code{FALSE}) which is faithful to the original description of
    the algorithm. This implementation is particularly useful for comparing the
    behaviour of different combinations of learning algorithms and statistical
    tests.
  \item a parallel implementation. It requires a running cluster set up with
    the \code{makeCluster} function from the \pkg{snow} package \citep{snow},
    which is passed to the function via the \code{cluster} argument.
\end{itemize}

The only available score-based learning algorithm is a \emph{Hill-Climbing}
(\code{hc}) greedy search on the space of directed graphs. The optimized
implementation (again used by default) uses score caching, score decomposability
and score equivalence to reduce the number of duplicated tests \citep{hcopt}.
Random restarts, a configurable number of perturbing operations and a preseeded
initial network structure can be used to avoid poor local maxima (with the
\code{restart}, \code{perturb} and \code{start} arguments, respectively).

\subsection{Conditional independence tests}

Several conditional independence tests from information theory and classical
statistics are available for use in constraint-based learning algorithms and
the \code{ci.test} function. In both cases the test to be used is specified
with the \code{test} argument (the label associated with each test is reported
in parenthesis).

Conditional independence tests for discrete data are functions of the conditional
probability tables implied by the graphical structure of the network through the
observed frequencies $\{n_{ijk}, i = 1, \ldots, R, j = 1, \ldots, C, k = 1, \ldots, L\}$
for the random variables $X$ and $Y$ and all the configurations of the conditioning
variables $\mathbf{Z}$:
\begin{itemize}
  \item \emph{mutual information}: an information-theoretic distance measure
    \citep{kullback}, defined as
    \begin{equation}
      \mathrm{MI}(X, Y \given \mathbf{Z}) = \sum_{i=1}^R \sum_{j=1}^C \sum_{k=1}^L
        \frac{n_{ijk}}{n} \log\frac{n_{ijk} n_{++k}}{n_{i+k} n_{+jk}}.
    \end{equation}
    It is proportional to the log-likelihood ratio test $\mathrm{G}^2$ (they
    differ by a $2n$ factor, where $n$ is the sample size) and it is related to
    the deviance of the tested models. Both the asymptotic $\chi^2$ test (\code{mi})
    and the Monte Carlo permutation test (\code{mc-mi}) described in \citet{good}
    are available.
  \item \emph{Pearson's $X^2$}: the classical Pearson's $X^2$ test for
    contingency tables,
    \begin{align}
      &X^2(X, Y \given \mathbf{Z}) = \sum_{i=1}^R \sum_{j=1}^C \sum_{k=1}^L
        \frac{\left(n_{ijk} - m_{ijk}\right)^2}{m_{ijk}},&
      &m_{ijk} = \frac{n_{i+k}n_{+jk}}{n_{++k}}
    \end{align}
    Again both the asymptotic $\chi^2$ test (\code{x2}) and a Monte Carlo 
    permutation test (\code{mc-x2}) from \citet{good} are available.
  \item \emph{fast mutual information} (\code{fmi}): a variant of the
    mutual information which is set to zero when there aren't at least
    five data per parameter, which is the usual threshold for establishing
    the correctness of the asymptotic $\chi^2$ distribution. This is the
    same heuristic defined for the Fast-IAMB algorithm in \citet{fastiamb2}.
  \item \emph{Akaike Information Criterion} (\code{aict}): an experimental
    AIC-based independence test, computed comparing the mutual information
    and the expected information gain. It rejects the null hypothesis if
    \begin{equation}
      \mathrm{MI}(X, Y \given \mathbf{Z}) \geqslant \frac{(R - 1)(C - 1)L}{n},
    \end{equation}
    which corresponds to an increase in the AIC score of the network.
\end{itemize}

In the continuous case conditional independence tests are functions of the
partial correlation coefficients $\rho_{XY \given \mathbf{Z}}$ of $X$ and
$Y$ given $\mathbf{Z}$:
\begin{itemize}
  \item \emph{linear correlation}: the linear correlation coefficient
    $\rho_{XY \given \mathbf{Z}}$.
    Both the asymptotic Student's $t$ test (\code{cor}) and the Monte Carlo
    permutation test (\code{mc-cor}) described in \citet{legendre}
    are available.
  \item \emph{Fisher's Z}: a transformation of the linear correlation coefficient
    used by commercial software (such as TETRAD) and the \pkg{pcalg} package 
    \citep{pcalg}, which implements the PC constraint-based learning algorithm
    \citep{spirtes}. It is defined as
    \begin{equation}
      \mathrm{Z}(X, Y \given \mathbf{Z}) = \frac{1}{2} \sqrt{n - |\mathbf{Z}| - 3}
        \log\frac{1 + \rho_{XY \given \mathbf{Z}}}{1 - \rho_{XY \given \mathbf{Z}}}.
    \end{equation}
    Both the asymptotic normal test
    (\code{zf}) and the Monte Carlo permutation test (\code{mc-zf})
    are available.
  \item \emph{mutual information} (\code{mi-g}): an information-theoretic
    distance measure \citep{kullback}, defined as
    \begin{equation}
      \mathrm{MI_g}(X, Y \given \mathbf{Z}) = - \frac{1}{2}
        \log(1 - \rho_{XY \given \mathbf{Z}}^2).
    \end{equation}
    It has the same relationship with the log-likelihood ratio as
    the corresponding test defined in the discrete case.
\end{itemize}

\subsection{Network scores}

Several score functions are available for use in the hill-climbing algorithm
and the \code{score} function. The score to be used is specified with the
\code{score} argument in \code{hc} and with the \code{type} argument in the
\code{score} function (the label associated with each score is reported in
parenthesis).

In the discrete case the following score functions are implemented:
\begin{itemize}
  \item the \emph{likelihood} (\code{lik}) and \emph{log-likelihood} (\code{loglik})
    scores, which are equivalent to the \emph{entropy measure} used by
    Weka \citep{weka}.
  \item the \emph{Akaike} (\code{aic}) and \emph{Bayesian} (\code{bic})
    \textit{Information Criterion} scores, defined as
    \begin{align}
      &\mathrm{AIC} = \log \mathrm{L}(X_1, \ldots, X_v) - d&
      &\mathrm{BIC} = \log \mathrm{L}(X_1, \ldots, X_v) - \frac{d}{2}\log n
    \end{align}
    The latter is equivalent to the \emph{Minimum Description Length} described by
    \citet{rissanen} and used as a Bayesian network score in \citet{mdl}.
  \item the logarithm of the \emph{Bayesian Dirichlet equivalent}
    score (\code{bde}), a score equivalent Dirichlet posterior density
    \citep{heckerman2}.
  \item the logarithm of the \emph{K2} score (\code{k2}), another Dirichlet
    posterior density \citep{k2} defined as
    \begin{align}
      &\mathrm{K2} = \prod_{i=1}^v \mathrm{K2}(X_i),&
       &\mathrm{K2}(X_i) = \prod_{j=1}^{L_i} \frac{(R_i - 1)!}{\left(\sum_{k=1}^{R_i} n_{ijk} + R_i - 1\right)!} \prod_{k=1}^{R_i} n_{ijk}!
    \end{align}
    and originally used in the structure learning algorithm of the same name.
    Unlike the \code{bde} score \code{k2} is not score equivalent.
\end{itemize}

The only score available for the continuous case is a score equivalent
{Gaussian posterior density} (\code{bge}), which follows a Wishart
distribution \citep{heckerman3}.

\subsection{Arc whitelisting and blacklisting}

Prior information on the data, such as the ones elicited from experts in the
relevant fields, can be integrated in all learning algorithms by means of the
\code{blacklist} and \code{whitelist} arguments. Both of them accept a set of
arcs which is guaranteed to be either present (for the former) or missing 
(for the latter) from the Bayesian network; any arc whitelisted and blacklisted
at the same time is assumed to be whitelisted, and is thus removed from the 
blacklist.

This combination represents a very flexible way to describe any arbitrary set
of assumptions on the data, and is also able to deal with partially directed
graphs:
\begin{itemize}
  \item any arc whitelisted in both directions (i.e. both $A \rightarrow B$
    and $B \rightarrow A$ are whitelisted) is present in the graph,
    but the choice of its direction is left to the learning algorithm.
    Therefore one of $A \rightarrow B$, $B \rightarrow A$ and $A - B$
    is guaranteed to be in the Bayesian network.
  \item any arc blacklisted in both directions, as well as the corresponding
    undirected arc, is never present in the graph. Therefore if both
    $A \rightarrow B$ and $B \rightarrow A$ are blacklisted, also $A - B$ is
    considered blacklisted.
  \item any arc whitelisted in one of its possible directions (i.e.
    $A \rightarrow B$ is whitelisted, but $B \rightarrow A$ is not)
    is guaranteed to be present in the graph in the specified direction.
    This effectively amounts to blacklisting both the corresponding
    undirected arc ($A - B$) and its reverse ($B \rightarrow A$).
  \item any arc blacklisted in one of its possible directions (i.e.
    $A \rightarrow B$ is blacklisted, but $B \rightarrow A$ is not)
    is never present in the graph. The same holds for $A - B$, but
    not for $B \rightarrow A$.
\end{itemize}

\section{A simple example}

In this section \pkg{bnlearn} will be used to analyze a small data set,
\code{learning.test}. It's included in the package itself along with other real
word and synthetic data sets, and is used in the example sections throughout
the manual pages due to its simple structure.

\subsection{Loading the package}

\pkg{bnlearn} and its dependencies (the \pkg{utils} package, which is bundled
with \proglang{R}) are available from CRAN, as are the suggested packages 
\pkg{snow} and \pkg{graph} \citep{graph}. The other suggested package, \pkg{Rgraphviz}
\citep{rgraphviz}, can be installed from BioConductor and is loaded along with
\pkg{bnlearn} if present.
\begin{Code}
> library(bnlearn)
Loading required package: Rgraphviz
Loading required package: graph
Loading required package: grid
Package Rgraphviz loaded successfully.
\end{Code}

\subsection{Learning a Bayesian network from data}

Once \pkg{bnlearn} is loaded, \code{learning.test} itself can be
loaded into a data frame of the same name with a call to \code{data}.
\begin{Code}
> data(learning.test)
> str(learning.test)
'data.frame':	5000 obs. of  6 variables:
 $ A: Factor w/ 3 levels "a","b","c": 2 2 1 1 1 3 3 2 2 2 ...
 $ B: Factor w/ 3 levels "a","b","c": 3 1 1 1 1 3 3 2 2 1 ...
 $ C: Factor w/ 3 levels "a","b","c": 2 3 1 1 2 1 2 1 2 2 ...
 $ D: Factor w/ 3 levels "a","b","c": 1 1 1 1 3 3 3 2 1 1 ...
 $ E: Factor w/ 3 levels "a","b","c": 2 2 1 2 1 3 3 2 3 1 ...
 $ F: Factor w/ 2 levels "a","b": 2 2 1 2 1 1 1 2 1 1 ...
\end{Code}
\code{learning.test} contains six discrete variables, stored as factors, each
with 2 (for \code{F}) or 3 (for \code{A}, \code{B}, \code{C}, \code{D} and
\code{E}) levels. The structure of the Bayesian network associated with this
data set can be learned for example with the Grow-Shrink algorithm, implemented
in the \code{gs} function, and stored in an object of class \code{bn}.
\begin{Code}
> bn.gs <- gs(learning.test)
> bn.gs

  Bayesian network learned via Constraint-based methods

  model:
    [partially directed graph] 
  nodes:                                 6 
  arcs:                                  5 
    undirected arcs:                     1 
    directed arcs:                       4 
  average markov blanket size:           2.33 
  average neighbourhood size:            1.67 
  average branching factor:              0.67 

  learning algorithm:                    Grow-Shrink 
  conditional independence test:         Mutual Information (discrete) 
  alpha threshold:                       0.05 
  tests used in the learning procedure:  43 
  optimized:                             TRUE 
\end{Code}

Other constraint-based algorithms return the same partially
directed network structure (again as an object of class \code{bn}),
as can be readily seen with \code{compare}.
\begin{Code}
> bn2 <- iamb(learning.test)
> bn3 <- fast.iamb(learning.test)
> bn4 <- inter.iamb(learning.test)
\end{Code}
\begin{Code}
> compare(bn.gs, bn2)
[1] TRUE
> compare(bn.gs, bn3)
[1] TRUE
> compare(bn.gs, bn4)
[1] TRUE
\end{Code}

On the other hand hill-climbing results in a completely directed network, which
differs from the previous one because the arc between \code{A} and \code{B} is
directed ($A \rightarrow B$ instead of $A - B$).
\begin{Code}
> bn.hc <- hc(learning.test, score = "aic")
> bn.hc

  Bayesian network learned via Score-based methods

  model:
    [A][C][F][B|A][D|A:C][E|B:F] 
  nodes:                                 6 
  arcs:                                  5 
    undirected arcs:                     0 
    directed arcs:                       5 
  average markov blanket size:           2.33 
  average neighbourhood size:            1.67 
  average branching factor:              0.83 

  learning algorithm:                    Hill-Climbing 
  score:                                 Akaike Information Criterion 
  penalization coefficient:              1 
  tests used in the learning procedure:  40 
  optimized:                             TRUE 

> compare(bn.hc, bn.gs)
[1] FALSE
\end{Code}
\begin{figure}[ht!]
  \begin{center}
    \includegraphics[angle=0,scale=1.25]{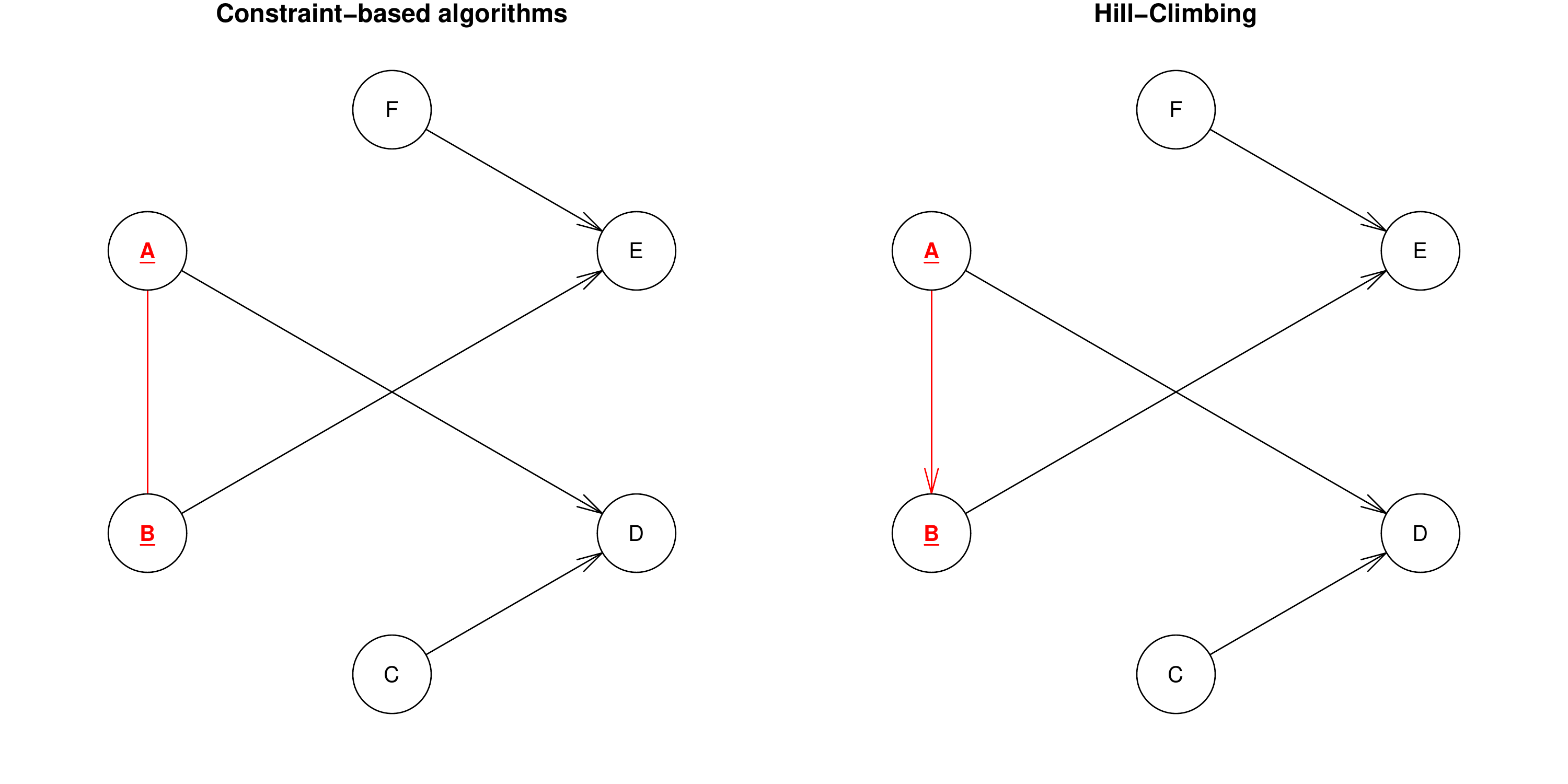}
    \caption{side by side comparison of the Bayesian network structures learned
      by constraint-based (\code{gs}, \code{iamb}, \code{fast.iamb} and
      \code{inter.iamb}) and score-based (\code{hc}) algorithms. Arcs which differ
      between the two network structures are plotted in red.}
    \label{fig:compare}
  \end{center}
\end{figure}
Another way to compare the two network structures is to plot them side by side
and highlight the differing arcs. This can be done either with the \code{plot} 
function (see Figure \ref{fig:compare}):
\begin{Code}
> par(mfrow = c(1,2))
> plot(bn.gs, main = "Constraint-based algorithms", highlight = c("A", "B"))
> plot(bn.hc, main = "Hill-Climbing", highlight = c("A", "B"))
\end{Code}
or with the more versatile \code{graphviz.plot}:
\begin{Code}
> par(mfrow = c(1,2))
> highlight.opts <- list(nodes = c("A", "B"), arcs = c("A", "B"),
+   col = "red", fill = "grey")
> graphviz.plot(bn.hc, highlight = highlight.opts)
> graphviz.plot(bn.gs, highlight = highlight.opts)
\end{Code}
which produces a better output for large graphs thanks to the functionality
provided by the \pkg{Rgraphviz} package.

The network structure learned by \code{gs}, \code{iamb}, \code{fast.iamb} and
\code{inter.iamb} is equivalent to the one learned by \code{hc}; changing the
arc $A - B$ to either $A \rightarrow B$ or to $B \rightarrow A$ results in 
networks with the same score because of the score equivalence property (which
holds for all the implemented score functions with the exception of $K2$).
Therefore if there is any prior information about the relationship between $A$
and $B$ the appropriate direction can be whitelisted (or its reverse can 
blacklisted, which is equivalent in this case).
\begin{Code}
> bn.AB <- gs(learning.test, blacklist = c("B", "A"))
> compare(bn.AB, bn.hc)
[1] TRUE
> score(bn.AB, learning.test, type = "bde")
[1] -24002.36
> bn.BA <- gs(learning.test, blacklist = c("A", "B"))
> score(bn.BA, learning.test, type = "bde")
[1] -24002.36
\end{Code}

\subsection{Network analysis and manipulation}

The structure of a Bayesian network is uniquely specified if its graph is
completely directed. In that case it can be represented as a string with
the \code{modelstring} function
\begin{Code}
> modelstring(bn.hc)
[1] "[A][C][F][B|A][D|A:C][E|B:F]"
\end{Code}
whose output is also included in the \code{print} method for the objects of
class \code{bn}. Each node is printed in square brackets along with all its
parents (which are reported after a pipe as a colon-separated list), and its
position in the string depends on the partial ordering defined by the network
structure. The same syntax is used in \pkg{deal} \citep{deal}, an \proglang{R}
package for learning Bayesian networks from mixed data. 

Partially directed graphs can be transformed into completely directed ones
with the \code{set.arc}, \code{drop.arc} and \code{reverse.arc} functions. For
example the direction of the arc $A - B$ in the \code{bn.gs} object can be set
to $A \rightarrow B$, so that the resulting network structure is identical to
the one learned by the hill-climbing algorithm.
\begin{Code}
> undirected.arcs(bn.gs)
     from to
[1,] "A"  "B"
[2,] "B"  "A"
> bn.dag <- set.arc(bn.gs, "A", "B")
> modelstring(bn.dag)
[1] "[A][C][F][B|A][D|A:C][E|B:F]"
> compare(bn.dag, bn.hc)
[1] TRUE
\end{Code}
Acyclicity is always preserved, as these commands return an error if the
requested changes would result in a cyclic graph.
\begin{Code}
> set.arc(bn.hc, "E", "A")
Error in arc.operations(x = x, from = from, to = to, op = "set",
check.cycles = check.cycles,  :
  the resulting graph contains cycles.
\end{Code}

Further information on the network structure can be extracted from
any \code{bn} object with the following functions:
\begin{itemize}
  \item whether the network structure is acyclic (\code{acyclic}) or
    completely directed (\code{directed});
  \item the labels of the nodes (\code{nodes}), of the root nodes (\code{root.nodes})
    and of the leaf nodes (\code{leaf.nodes});
  \item the directed arcs (\code{directed.arcs}) of the network, the undirected ones
    (\code{undirected.arcs}) or both of them (\code{arcs});
  \item the adjacency matrix (\code{amat}) and the number of parameters (\code{nparams})
    associated with the network structure;
  \item the parents (\code{parents}), children (\code{children}), Markov blanket
    (\code{mb}), and neighbourhood (\code{nbr}) of each node.
\end{itemize}

The \code{arcs}, \code{amat} and \code{modelstring} functions can also be
used in combination with \code{empty.graph} to create a \code{bn} object
with a specific structure from scratch:
\begin{Code}
> other <- empty.graph(nodes = nodes(bn.hc))
\end{Code}
\begin{Code}
> arcs(other) <- data.frame(
+   from = c("A", "A", "B", "D"),
+   to = c("E", "F", "C", "E"))
> other

  Randomly generated Bayesian network

  model:
    [A][B][D][C|B][E|A:D][F|A] 
  nodes:                                 6 
  arcs:                                  4 
    undirected arcs:                     0 
    directed arcs:                       4 
  average markov blanket size:           1.67 
  average neighbourhood size:            1.33 
  average branching factor:              0.67 

  generation algorithm:                  Empty 

\end{Code}
This is particularly useful to compare different network structures for the same
data, for example to verify the goodness of fit of the learned network with
respect to a particular score function.
\begin{Code}
> score(other, data = learning.test, type = "aic")
[1] -28019.79
> score(bn.hc, data = learning.test, type = "aic")
[1] -23873.13
\end{Code}

\subsection{Debugging utilities and diagnostics}

Many of the functions of the \pkg{bnlearn} package are able to print additional
diagnostic messages if called with the \code{debug} argument set to \code{TRUE}.
This is especially useful to study the behaviour of the learning algorithms in
specific settings and to investigate anomalies in their results (which may be
due to an insufficient sample size for the asymptotic distribution of the tests
to be valid, for example).
For example the debugging output of the call to \code{gs} previously used to 
produce the \code{bn.gs} object reports the exact sequence of conditional 
independence tests performed by the learning algorithm, along with the effects
of the backtracking optimizations (some parts are omitted for brevity).

\begin{Code}
> gs(learning.test, debug = TRUE)
----------------------------------------------------------------
* learning markov blanket of A .
  * checking node B for inclusion.
    > node B included in the markov blanket ( p-value: 0 ).
    > markov blanket now is ' B '.
  * checking node C for inclusion.
    > A indep. C given ' B ' ( p-value: 0.8743202 ).
  * checking node D for inclusion.
    > node D included in the markov blanket ( p-value: 0 ).
    > markov blanket now is ' B D '.
  * checking node E for inclusion.
    > A indep. E given ' B D ' ( p-value: 0.5193303 ).
  * checking node F for inclusion.
    > A indep. F given ' B D ' ( p-value: 0.07368042 ).
  * checking node C for inclusion.
    > node C included in the markov blanket ( p-value: 1.023194e-254 ).
    > markov blanket now is ' B D C '.
  * checking node E for inclusion.
    > A indep. E given ' B D C ' ( p-value: 0.5091863 ).
  * checking node F for inclusion.
    > A indep. F given ' B D C ' ( p-value: 0.3318902 ).
  * checking node B for exclusion (shrinking phase).
    > node B remains in the markov blanket. ( p-value: 9.224694e-291 )
  * checking node D for exclusion (shrinking phase).
    > node D remains in the markov blanket. ( p-value: 0 )
----------------------------------------------------------------
* learning markov blanket of B .
[...]
----------------------------------------------------------------
* learning markov blanket of F .
    * known good (backtracking): ' B E '.
    * known bad (backtracking): ' A C D '.
    * nodes still to be tested for inclusion: '  '.
----------------------------------------------------------------
* checking consistency of markov blankets.
[...]
----------------------------------------------------------------
* learning neighbourhood of A .
  * blacklisted nodes: '  '
  * whitelisted nodes: '  '
  * starting with neighbourhood: ' B D C '
  * checking node B for neighbourhood.
    > dsep.set = ' E F '
    > trying conditioning subset '  '.
    > node B is still a neighbour of A . ( p-value: 0 )
    > trying conditioning subset ' E '.
    > node B is still a neighbour of A . ( p-value: 0 )
    > trying conditioning subset ' F '.
    > node B is still a neighbour of A . ( p-value: 0 )
    > trying conditioning subset ' E F '.
    > node B is still a neighbour of A . ( p-value: 0 )
  * checking node D for neighbourhood.
    > dsep.set = ' C '
    > trying conditioning subset '  '.
    > node D is still a neighbour of A . ( p-value: 0 )
    > trying conditioning subset ' C '.
    > node D is still a neighbour of A . ( p-value: 0 )
  * checking node C for neighbourhood.
    > dsep.set = ' D '
    > trying conditioning subset '  '.
    > node C is not a neighbour of A . ( p-value: 0.8598334 )
----------------------------------------------------------------
* learning neighbourhood of B .
[...]
----------------------------------------------------------------
* learning neighbourhood of F .
  * blacklisted nodes: '  '
  * whitelisted nodes: '  '
  * starting with neighbourhood: ' E '
  * known good (backtracking): ' E '.
  * known bad (backtracking): ' A B C D '.
----------------------------------------------------------------
* checking consistency of neighbourhood sets.
[...]
----------------------------------------------------------------
* v-structures centered on D .
  * checking A -> D <- C
    > chosen d-separating set: '  '
    > testing A vs C given  D ( 0 )
    @ detected v-structure A -> D <- C
----------------------------------------------------------------
* v-structures centered on E .
  * checking B -> E <- F
    > chosen d-separating set: '  '
    > testing B vs F given  E ( 1.354269e-50 )
    @ detected v-structure B -> E <- F
----------------------------------------------------------------
* v-structures centered on F .
----------------------------------------------------------------
* applying v-structure A -> D <- C ( 0.000000e+00 )
* applying v-structure B -> E <- F ( 1.354269e-50 )
----------------------------------------------------------------
* detecting cycles ...
[...]
----------------------------------------------------------------
* propagating directions for the following undirected arcs:
[...]
\end{Code}
Other functions which provide useful diagnostics include (but are not limited
to) \code{compare} (which reports the differences between the two network
structures with respect to arcs, parents and children for each node), \code{score}
and \code{nparams} (which report the number of parameters and the
contribution of each node to the network score, respectively).

\section{Practical examples}

\subsection{The ALARM network}

The ALARM (``A Logical Alarm Reduction Mechanism'') network from \citet{alarm}
is a Bayesian network designed to provide an alarm message system for patient
monitoring. It has been widely used (see for example \citet{mmhc} and \citet{sc})
as a case study to evaluate the performance of new structure learning algorithms.

\begin{figure}[t!]
  \begin{center}
    \includegraphics[angle=0,scale=1.25]{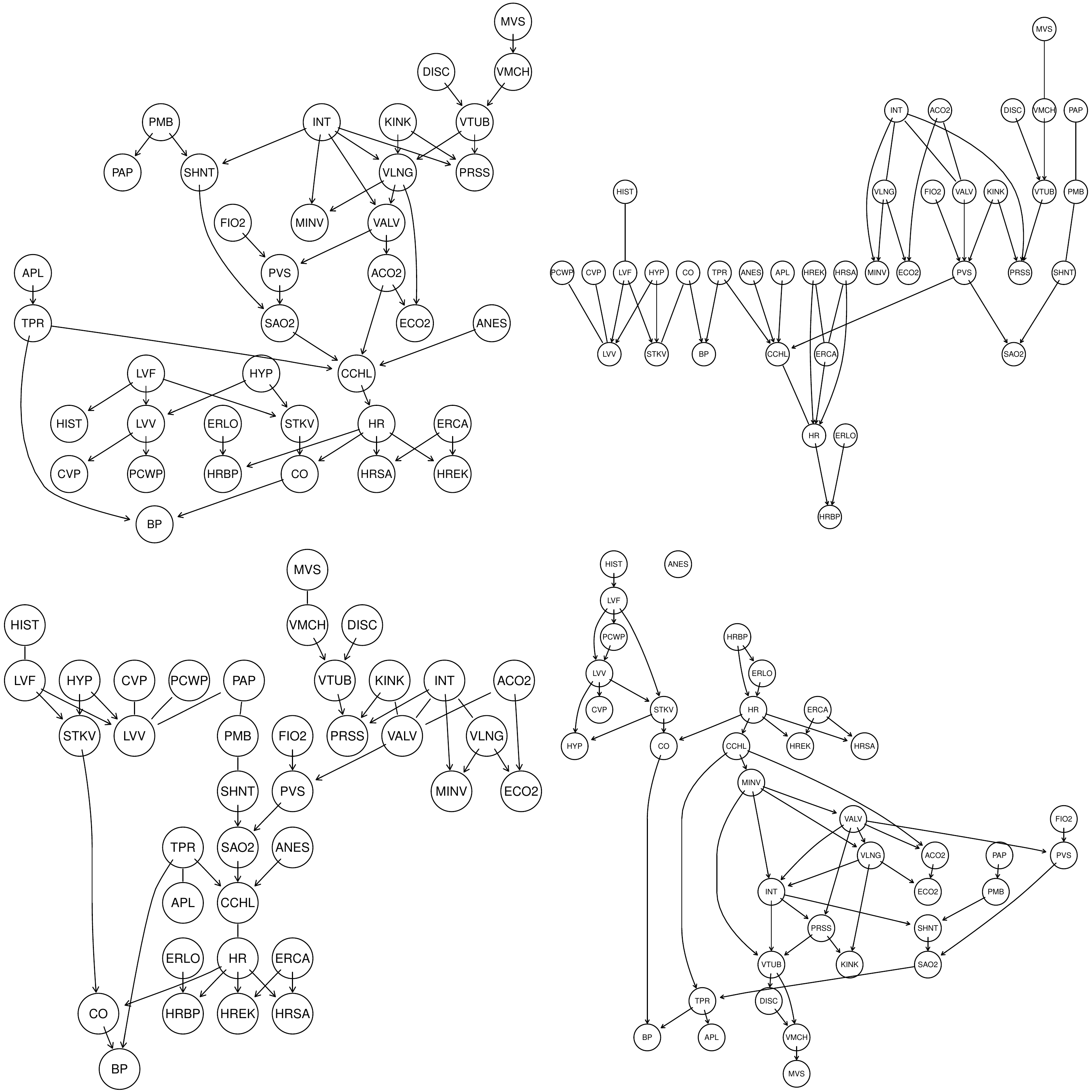}
    \caption{The ALARM data set: the original network structure (top left)
    and the network structures learned by \code{gs} (top right),
    \code{inter.iamb} (bottom left) and \code{hc} (bottom right).}
    \label{fig:alarm}
  \end{center}
\end{figure}

The \code{alarm} data set includes a sample of size 20000 generated from
this network, which contains 37 discrete variables (with two to four levels
each) and 46 arcs. Every learning algorithm implemented in \pkg{bnlearn} (except
\code{mmpc}) is capable of recovering the ALARM network to within a few arcs
and arc directions (see Figure \ref{fig:alarm}).
\begin{Code}
> alarm.gs <- gs(alarm)
> alarm.iamb <- iamb(alarm)
> alarm.fast.iamb <- fast.iamb(alarm)
> alarm.inter.iamb <- inter.iamb(alarm)
> alarm.mmpc <- mmpc(alarm)
> alarm.hc <- hc(alarm, score = "bic")
\end{Code}
The number of conditional independence tests, which provides an implementation
independent performance indicator, is similar in all constraint-based algorithms
(see Table \ref{table:alarm}); the same holds for the number of network score
comparisons performed by \code{hc}, even though the learned network has about
ten more arcs than the other ones.
\begin{table}[b!]
  \begin{center}
    \begin{tabular}{|l|c|c|c|c|c|c|}
    \hline
                               & \code{gs}  & \code{iamb} & \code{fast.iamb} & \code{inter.iamb} & \code{hc} \\
    \hline
    independence tests /       & 1727       & 2874        & 2398             & 3106              & 2841      \\
    network comparisons        &            &             &                  &                   &           \\
    \hline
    learned arcs               & 42         & 43          & 45               & 43                & 53        \\
    (directed/undirected)      & (29/13)    & (29/14)     & (32/13)          & (30/13)           & (53/0)    \\
    \hline
    execution time             & 13.54360   & 17.63735    & 14.80890         & 18.59825          & 72.38705  \\
    \hline
    \end{tabular}
    \caption{Performance of implemented learning algorithms with the \code{alarm}
      data set, measured in the number of conditional independence tests (for
      constraint-based algorithms) or network score comparisons (for score-based
      algorithms), the number of arcs and the execution time on an Intel Core
      2 Duo machine with 1GB of RAM.}
    \label{table:alarm}
  \end{center}
\end{table}

The quality of the learned network improves significantly if the
Monte Carlo versions of the tests are used instead of the parametric
ones, as probability structure of the ALARM network results in many
sparse contingency tables. For example a side by side comparison of
the two versions of Pearson's $X^2$ test shows that the use of the
nonparametric tests leads to the correct identification of all but
five arcs, instead of the 12 missed with the parametric tests.
\begin{Code}
> dag <- empty.graph(names(alarm))
> modelstring(dag) <- paste("[HIST|LVF][CVP|LVV][PCWP|LVV][HYP][LVV|HYP:LVF]",
+  "[LVF][STKV|HYP:LVF][ERLO][HRBP|ERLO:HR][HREK|ERCA:HR][ERCA][HRSA|ERCA:HR]",
+  "[ANES][APL][TPR|APL][ECO2|ACO2:VLNG][KINK][MINV|INT:VLNG][FIO2]",
+  "[PVS|FIO2:VALV][SAO2|PVS:SHNT][PAP|PMB][PMB][SHNT|INT:PMB][INT]",
+  "[PRSS|INT:KINK:VTUB][DISC][MVS][VMCH|MVS][VTUB|DISC:VMCH]",
+  "[VLNG|INT:KINK:VTUB][VALV|INT:VLNG][ACO2|VALV][CCHL|ACO2:ANES:SAO2:TPR]",
+  "[HR|CCHL][CO|HR:STKV][BP|CO:TPR]", sep = "")
> alarm.gs <- gs(alarm, test = "x2")
> alarm.mc <- gs(alarm, test = "mc-x2", B = 10000)
> par(mfrow = c(1,2), omi = rep(0, 4), mar = c(1, 0, 1, 0))
> graphviz.plot(dag, highlight = list(arcs = arcs(alarm.gs)))
> graphviz.plot(dag, highlight = list(arcs = arcs(alarm.mc)))
\end{Code}
\begin{figure}[t!]
  \begin{center}
    \includegraphics[angle=0,scale=1.25]{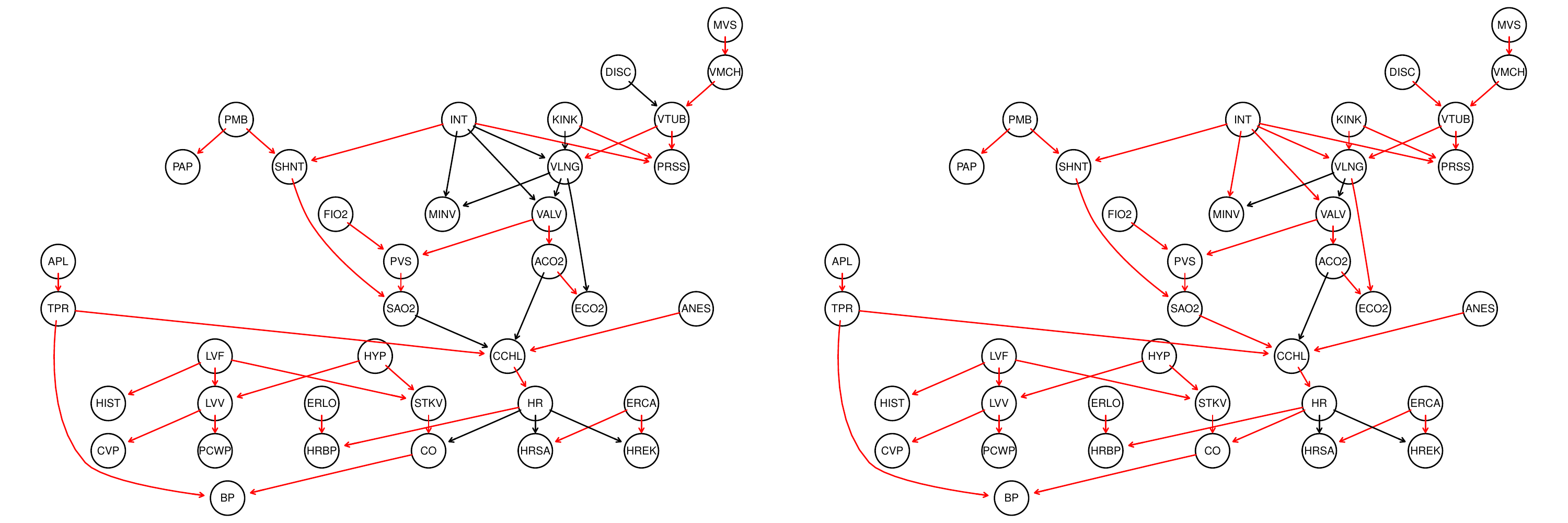}
    \caption{Side by side comparison of the network structures learned from
      the \code{alarm} data set by the Grow-Shrink algorithm with the
      parametric (on the left) and nonparametric (on the right) versions
      of Pearson's $X^2$ test. The arcs of the true network structure
      present in each case are highlighted in red.}
    \label{fig:alarm2}
  \end{center}
\end{figure}

\subsection{The examination marks data set}

The \code{marks} data set is a small data set studied in \citet{mardia}, \citet{whittaker}
and \citet{edwards}. It contains five continuous variables, the examination
marks for 88 students in five different subjects (mechanics, vectors, algebra,
analysis and statistics).
\begin{Code}
> data(marks)
> str(marks)
'data.frame':	88 obs. of  5 variables:
 $ MECH: num  77 63 75 55 63 53 51 59 62 64 ...
 $ VECT: num  82 78 73 72 63 61 67 70 60 72 ...
 $ ALG : num  67 80 71 63 65 72 65 68 58 60 ...
 $ ANL : num  67 70 66 70 70 64 65 62 62 62 ...
 $ STAT: num  81 81 81 68 63 73 68 56 70 45 ...
\end{Code}
The purpose of the analysis was to find a suitable way to combine or average
these marks. Since they are obviously correlated, the exact weights they are
assigned depend on the estimated dependence structure of the data.

Under the assumption of multivariate normality this analysis requires the
examination of the partial correlation coefficients, some of which are clearly not
significative:
\begin{Code}
> ci.test("MECH", "ANL", "ALG", data = marks)

	Pearson's Linear Correlation

data:  MECH ~ ANL | ALG
cor = 0.0352, df = 85, p-value = 0.7459
alternative hypothesis: true value is not equal to 0

> ci.test("STAT", "VECT", "ALG", data = marks)

	Pearson's Linear Correlation

data:  STAT ~ VECT | ALG
cor = 0.0527, df = 85, p-value = 0.628
alternative hypothesis: true value is not equal to 0
\end{Code}
\begin{figure}[t!]
  \begin{center}
    \hspace{-0.8cm}
    \includegraphics[angle=0,scale=1.3]{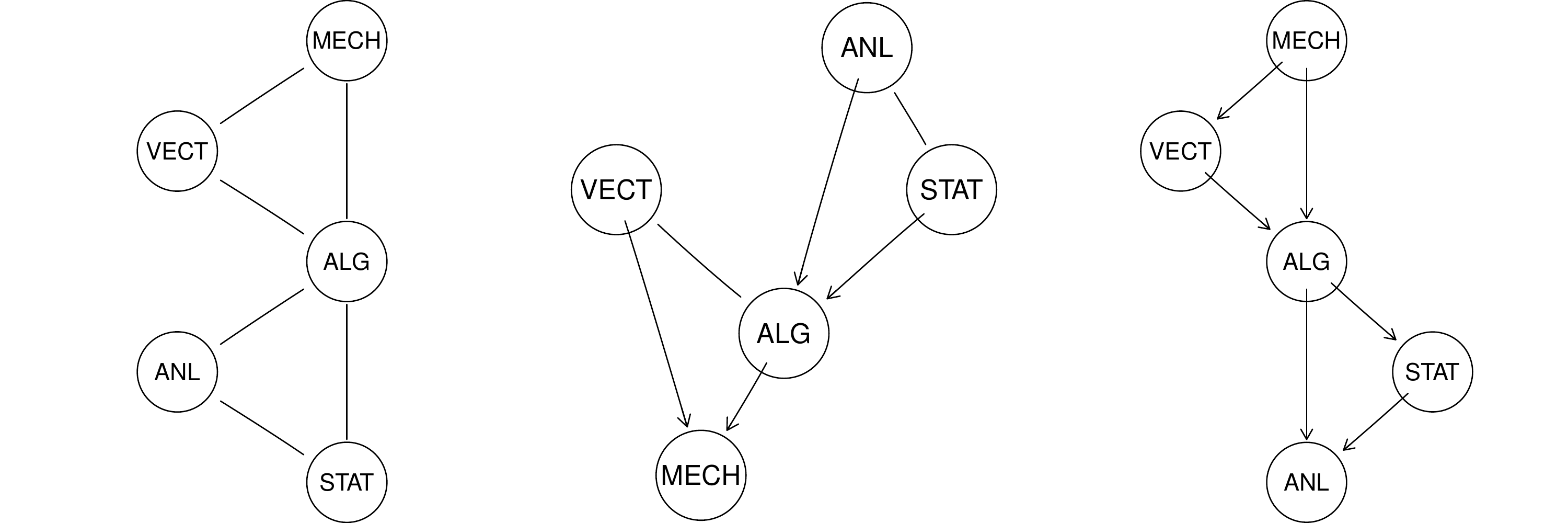}
    \caption{Graphical models learned from the marks data set. From left to
      right: the network learned by \code{mmpc} (which is identical to the
      Gaussian graphical model in \citet{edwards}), the one learned by the
      other constraint-based algorithms and the one learned by \code{hc}.}
    \label{fig:marks}
  \end{center}
\end{figure}
This is confirmed by the other conditional independence tests, both
parametric and nonparametric; for example:
\begin{Code}
> ci.test("STAT", "VECT", "ALG", data = marks, test = "zf")$p.value
[1] 0.6289112
> ci.test("STAT", "VECT", "ALG", data = marks, test = "mc-cor")$p.value
[1] 0.6332
> ci.test("STAT", "VECT", "ALG", data = marks, test = "mi-g")$p.value
[1] 0.6209406
> ci.test("STAT", "VECT", "ALG", data = marks, test = "mc-mi-g")$p.value
[1] 0.6226
\end{Code}

All learning algorithms result in very similar network structures, which agree
up to the direction of the arcs (see Figure \ref{fig:marks}). In all models the
marks for analysis and statistics are conditionally independent from the ones
for mechanics and vectors, given algebra. The structure of the graph suggests
that the latter is essential in the overall evaluation of the examination.

\section{Other packages for learning Bayesian networks}

There exist other packages in \proglang{R} which are able to either learn the 
structure of a Bayesian network or fit and manipulate its parameters. Some
examples are \pkg{pcalg}, which implements the PC algorithm and focuses on the
causal interpretation of Bayesian networks; \pkg{deal}, which implements a 
hill-climbing search for mixed data; and the suite composed by \pkg{gRbase}
\citep{grbase}, \pkg{gRain} \citep{grain}, \pkg{gRc} \citep{grc}, which 
implements various exact and approximate inference procedures.

However, none of these packages is as versatile as \pkg{bnlearn} for learning
the structure of Bayesian networks. \pkg{deal} and \pkg{pcalg} implement a 
single learning algorithm, even though are able to handle both discrete and
continuous data. Furthermore, the PC algorithm has a poor performance in terms
of speed and accuracy compared to newer constraint-based algorithms such as
Grow-Shrink and IAMB \citep{iamb}. \pkg{bnlearn} also offers a wider selection
of network scores and conditional independence tests; in particular it's the
only \proglang{R} package able to learn the structure of Bayesian networks
using permutation tests, which are superior to the corresponding asymptotic
tests at low sample sizes.

\section{Conclusions}

\pkg{bnlearn} is an \proglang{R} package which provides a free implementation of
some Bayesian network structure learning algorithms appeared in recent literature,
enhanced with algorithmic optimizations and support for parallel computing.
Many score functions and conditional independence tests are provided for both 
independent use and the learning algorithms themselves.

\pkg{bnlearn} is designed to provide the versatility needed to handle experimental
data analysis. It handles both discrete and continuous data, and it supports any
combination of the implemented learning algorithms and either network scores (for
score-based algorithms) or conditional independence tests (for constraints-based
algorithms). Furthermore, it simplifies the analysis of the learned networks by
providing a single object class (\code{bn}) for all the algorithms and a set of
utility functions to perform descriptive statistics and basic inference procedures.

\section*{Acknowledgements}

Many thanks to Prof. Adriana Brogini, my Supervisor at the Ph.D. School in Statistical
Sciences (University of Padova), for proofreading this article and giving many
useful comments and suggestions. I would also like to thank Radhakrishnan Nagarajan
(University of Arkansas for Medical Sciences) and Suhaila Zainudin (Universiti 
Teknologi Malaysia) for their support, which encouraged me in the development
of this package.

\end{document}